\pdfoutput=1
\documentclass[letterpaper]{article} % DO NOT CHANGE THIS
\usepackage{aaai2026}  % DO NOT CHANGE THIS
\usepackage{times}  % DO NOT CHANGE THIS
\usepackage{helvet}  % DO NOT CHANGE THIS
\usepackage{courier}  % DO NOT CHANGE THIS
\usepackage[hyphens]{url}  % DO NOT CHANGE THIS
\usepackage{graphicx} % DO NOT CHANGE THIS
\usepackage{natbib}  % DO NOT CHANGE THIS AND DO NOT ADD ANY OPTIONS TO IT
\usepackage{caption} % DO NOT CHANGE THIS AND DO NOT ADD ANY OPTIONS TO IT
\usepackage{algorithm}
\usepackage{algorithmic}
\usepackage{comment}
\usepackage{soul}
\usepackage{amssymb}
\usepackage{amsmath}
\usepackage{multirow}
\usepackage{newfloat}
\usepackage{listings}
\DeclareCaptionStyle{ruled}{labelfont=normalfont,labelsep=colon,strut=off} % DO NOT CHANGE THIS
\floatstyle{ruled}
\newfloat{listing}{tb}{lst}{}
\floatname{listing}{Listing}
\title{AI-Driven Evaluation of Surgical Skill via Action Recognition}
\author{
    Yan Meng\textsuperscript{\rm 1}\thanks{ymeng@childrensnational.org}, Daniel Donoho\textsuperscript{\rm 1},  Marcelle Altshuler\textsuperscript{\rm 2}, Omar Arnaout\textsuperscript{\rm 2}
}
\affiliations{
    \textsuperscript{\rm 1}Children's National Hospital\\
    111 Michigan Ave NW,\\
    Washington, DC 20010 USA\\
	\textsuperscript{\rm 2}Brigham and Women’s Hospital, Harvard Medical School\\
	Boston, MA 02115 USA\\
}

\usepackage{bibentry}
\begin{document}

\maketitle

\begin{abstract}
Mastery of microanastomosis represents a fundamental competency in neurosurgery, where the ability to perform highly precise and coordinated movements under a microscope is directly correlated with surgical success and patient safety. These procedures demand not only fine motor skills but also sustained concentration, spatial awareness, and dexterous bimanual coordination. As such, the development of effective training and evaluation strategies is critical. Conventional methods for assessing surgical proficiency typically rely on expert supervision, either through onsite observation or retrospective analysis of recorded procedures. However, these approaches are inherently subjective, susceptible to inter-rater variability, and require substantial time and effort from expert surgeons. These demands are often impractical in low- and middle-income countries, thereby limiting the scalability and consistency of such methods across training programs. To address these limitations, we propose a novel AI-driven framework for the automated assessment of microanastomosis performance. The system integrates a video transformer architecture based on TimeSformer, improved with hierarchical temporal attention and weighted spatial attention mechanisms, to achieve accurate action recognition within surgical videos. Fine-grained motion features are then extracted using a YOLO-based object detection and tracking method, allowing for detailed analysis of instrument kinematics. Performance is evaluated along five aspects of microanastomosis skill, including overall action execution, motion quality during procedure-critical actions, and general instrument handling. Experimental validation using a dataset of 58 expert-annotated videos demonstrates the effectiveness of the system, achieving 87.7\% frame-level accuracy in action segmentation that increased to 93.62\% with post-processing, and an average classification accuracy of 76\% in replicating expert assessments across all skill aspects. These findings highlight the system’s potential to provide objective, consistent, and interpretable feedback, thereby enabling more standardized, data-driven training and evaluation in surgical education.

\end{abstract}


\section{Introduction}

Accurate and consistent assessment of technical skill remains a longstanding challenge in surgical education. This is particularly evident in microsurgical procedures such as microanastomosis, where surgeons are required to manipulate submillimeter-scale vessels under high magnification. A typical end-to-side microanastomosis procedure involves cutting the donor vessel and suturing it to the recipient vessel using eight evenly spaced stitches positioned circumferentially around the arteriotomy site. The ability to segment and analyze such fine-grained actions is critical for skill acquisition, performance evaluation, and long-term quality assurance in both training and clinical practice.

Traditional approaches for surgical skill assessment rely heavily on expert raters who manually review procedural videos using structured rubrics, such as the Objective Structured Assessment of Technical Skill (OSATS) \cite{martin1997objective}, Global Rating Scales (GRS) \cite{regehr1998comparing}, and Neurosurgical Objective Microanastomosis Assessment Tool (NOMAT) \cite{aoun2015pilot}. While these methods are valuable, they are labor-intensive, inherently subjective, and difficult to scale. Recent advancements have attempted to address these limitations through the use of sensor data, motion tracking, or robotic kinematic analysis. However, these methods typically require specialized equipment and may lack interpretability, particularly when assessing discrete, task-specific gestures \cite{zia2016automated,funke2019video,lavanchy2021automation,meng2023automatic}.

In parallel, the computer vision community has made substantial progress in video understanding, particularly in domains requiring fine temporal resolution and semantic interpretability. Transformer-based architectures, in particular, have emerged as a powerful class of models for sequential data, owing to their capacity to model long-range dependencies via self-attention mechanisms \cite{bertasius2021space, mazzia2022action, wang2022efficient,yang2024surgformer}. In the context of video analysis, these models have achieved state-of-the-art performance in action recognition. Their ability to incorporate temporal context makes them well-suited for analyzing surgical workflows, where actions may be subtle and semantically interdependent.

To advance the field of automated surgical skill assessment, we propose a novel AI framework designed to perform interpretable, action-level evaluation of microanastomosis performance. Our system consists of three primary components: (1) a low-cost, self-guided microanastomosis training kit to facilitate self-paced practice and data collection; (2) a transformer-based video segmentation module to automatically identify pre-defined surgical actions; and (3) a YOLO-based instrument detection and tracking pipeline to extract motion features within each segmented action.

By integrating spatial and temporal context, our approach enables the extraction of interpretable metrics such as action duration, repetition, and kinematic patterns aligned with specific surgical tasks. These features are subsequently used in a supervised learning framework to replicate expert NOMAT scoring and provide fine-grained, objective feedback.

Beyond methodological contributions, this work has broader implications for global health and surgical education. According to the Lancet Commission on Global Surgery, an estimated 5 billion people lack access to safe and timely surgical care, with low- and middle-income countries (LMICs) disproportionately affected due to resource constraints and limited access to specialist training \cite{meara2015global}. In such contexts, scalable and interpretable assessment tools based on surgical video offer a cost-effective pathway to enhance skill acquisition, reduce variability in training, and expand access to high-quality surgical education. By democratizing technical feedback through automation, this research supports a more equitable and data-driven approach to global surgical capacity building.

Our main contributions are summarized as follows:
\begin{itemize}
	\item We present an AI framework for automated, action-level skill assessment in microanastomosis procedures, offering interpretable and task-specific feedback.
	\item We integrate transformer-based action segmentation with YOLO-based instrument detection and tracking to extract kinematic features aligned with surgical actions.
	\item We demonstrate the feasibility of using supervised classification to replicate expert NOMAT scores at the coarse and fine action level.
	\item We highlight the broader impact of our framework in addressing disparities in surgical training, particularly in resource-limited settings.
\end{itemize}

\section{Related Work}

\subsubsection{Video-Based Surgical Skill Assessment.}
Traditional surgical assessment frameworks such as the OSATS, GRS, and NOMAT offer expert-based evaluations guidelines. However, human rating are subjective, time-intensive, and difficult to scale. To address these limitations, the automation of surgical skill assessment using video data has been explored across multiple studies, often combining spatial and temporal modeling with domain-specific metrics. Early work employed convolutional Neural Network (CNN ) and motion feature pipelines to classify surgeon expertise using videos, achieving high accuracy in binary scenario \cite{funke2019video}. Similarly, SATR‑DL\cite{wang2018satr} performed end-to-end task and skill recognition using motion profile analysis gleaned from robot-assisted surgery data, achieving excellent accuracy in distinguishing expertise levels; and a JAMA Network Open study \cite{kitaguchi2021development} generalize the 3D CNN-based spatiotemporal modeling across diverse intraoperative videos, attaining acceptable accuracy in categorizing surgical steps based on expert ratings. 

More recent systems integrate attention and auxiliary supervision. For instance, ViSA \cite{li2022surgical} models heterogeneous semantic parts and aggregates them temporally to assess skill, enhancing interpretability and performance. A VBA‑Net framework\cite{yanik2023video} provide both formative and summative skill assessment using attention heatmaps to highlight formative feedback. Other studies based on hand-crafted motion metrics such as path length, number of movements, instrument orientation extracted from bounding boxes have shown that video analysis yield statistically significant discrimination between skill levels in laparoscopic tasks\cite{goldbraikh2022video, hung2023capturing}.

\begin{figure*}[!th]
	\centering
	\includegraphics[width=\linewidth]{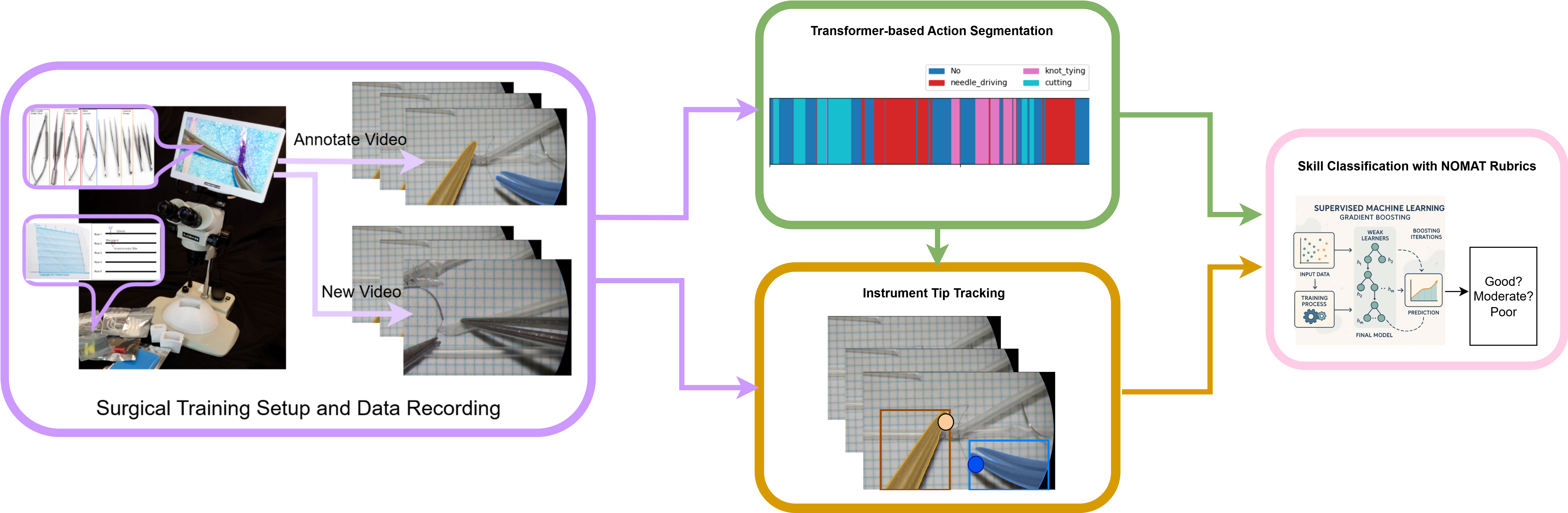}
	\caption{An overview of the transformer-based microanastomosis skill assessment framework.}
	\label{fig:framework}
\end{figure*}

\subsubsection{Temporal Action Segmentation.}
In recent years, there has been significant progress in temporal convolutional networks (TCNs) and transformer-based architectures for video understanding tasks, particularly in predicting action classes at fine temporal resolutions across video sequences.

Temporal convolutional models, such as MS-TCN \cite{farha2019ms} and TECNO \cite{czempiel2020tecno}, have been widely adopted for action segmentation. These models leverage successive convolutional stages to progressively refine per-frame action predictions while capturing hierarchical temporal context. Despite their effectiveness, TCNs can struggle to model long-range dependencies and may be less suitable for complex procedural tasks due to their limited receptive fields.

Transformer-based architectures have demonstrated superior performance across various video understanding benchmarks. A notable early example is TimeSformer \cite{bertasius2021space}, which introduced a factorized self-attention mechanism alternating between spatial and temporal dimensions. This architecture achieved state-of-the-art results on large-scale datasets of short videos such as Kinetics-400 and Kinetics-600, showcasing its strength in temporal modeling compared to conventional 3D convolutional networks. Building on this foundation, ASFormer \cite{yi2021asformer} improved temporal segmentation by incorporating explicit local connectivity priors, hierarchical representations to manage extended input sequences, and a decoder designed to refine coarse action predictions. ASFormer achieved strong performance on standard action segmentation benchmarks, particularly under limited data conditions. Similarly, ActionFormer \cite{zhang2022actionformer} proposed an efficient transformer-based architecture for temporal action localization. By combining multi-scale feature representations, localized self-attention, and a lightweight decoding module, ActionFormer outperformed previous methods such as SlowFast and I3D. Hybrid approaches, such as ASTCFormer \cite{zhang2023surgical}, have further integrated TCNs with transformer layers to jointly capture local temporal continuity and global dependencies in surgical workflow recognition tasks. However, these increasingly complex architectures do not always translate into substantial performance gains.

While several existing models address surgical skill assessment using video or kinematic data, few provide interpretable, action-level evaluations aligned with established scoring rubrics. Most approaches were validated only on short videos or lack fine-grained temporal segmentation. Moreover, the application of transformer-based temporal segmentation in surgical video analysis remains relatively underexplored. Positioned at this intersection, our work leverages and extends the TimeSformer architecture for frame-level segmentation of predefined surgical actions. Augmented with YOLO-based surgical instrument tip localization and tracking, our pipeline enables per-action kinematic analysis and alignment with expert rubrics. This facilitates interpretable, action-level skill classification and narrows the gap between the domain-specific demands of surgical education and the capabilities of state-of-the-art video understanding models.

\section{Methodology}
\begin{comment}
	We propose a transformer-based framework for action-level microanastomosis skill assessment. The framework comprises four main components: a data recording module; a transformer-based action segmentation module; a YOLO-based instrument kinematics feature extraction module; and a supervised skill classification module guided by the NOMAT rubrics. The complete pipeline is illustrated in Figure X.
\end{comment}
We propose a transformer-based framework for action-level assessment of microanastomosis skills. The framework comprises (i) a self-guided microanastomosis practice kit with an integrated data recording component; (ii) a transformer-based action segmentation module; (iii) a YOLO-based instrument kinematics feature extraction module; and (iv) a supervised skill classification module guided by the NOMAT rubric. An overview of the complete pipeline is presented in Fig. \ref{fig:framework}.

\subsection{Self-guided Microanastomosis Practice Kit}
\label{sec:kit}
The microanastomosis training toolkit used in this study is designed to provide a standardized and replicable environment for simulating small-caliber vascular procedures. Each setup includes a Meiji Techno EMZ-250TR trinocular zoom stereomicroscope, paired with a high-definition camera and external monitor to facilitate visual observation and high-quality data capture. Vascular tissues are simulated using 1.0mm $\times$ 0.8mm microvascular practice cards from Pocket Suture, USA, designed to replicate the size and mechanical behavior of small-caliber vessels. A uniform set of microsurgical instruments is provided with each kit to ensure procedural consistency across users. The instrument set comprised one straight needle driver, one curved needle driver, and a pair of both straight and curved microsurgical scissors. 

\begin{figure}[!t]
	\centering
	\includegraphics[width=\columnwidth]{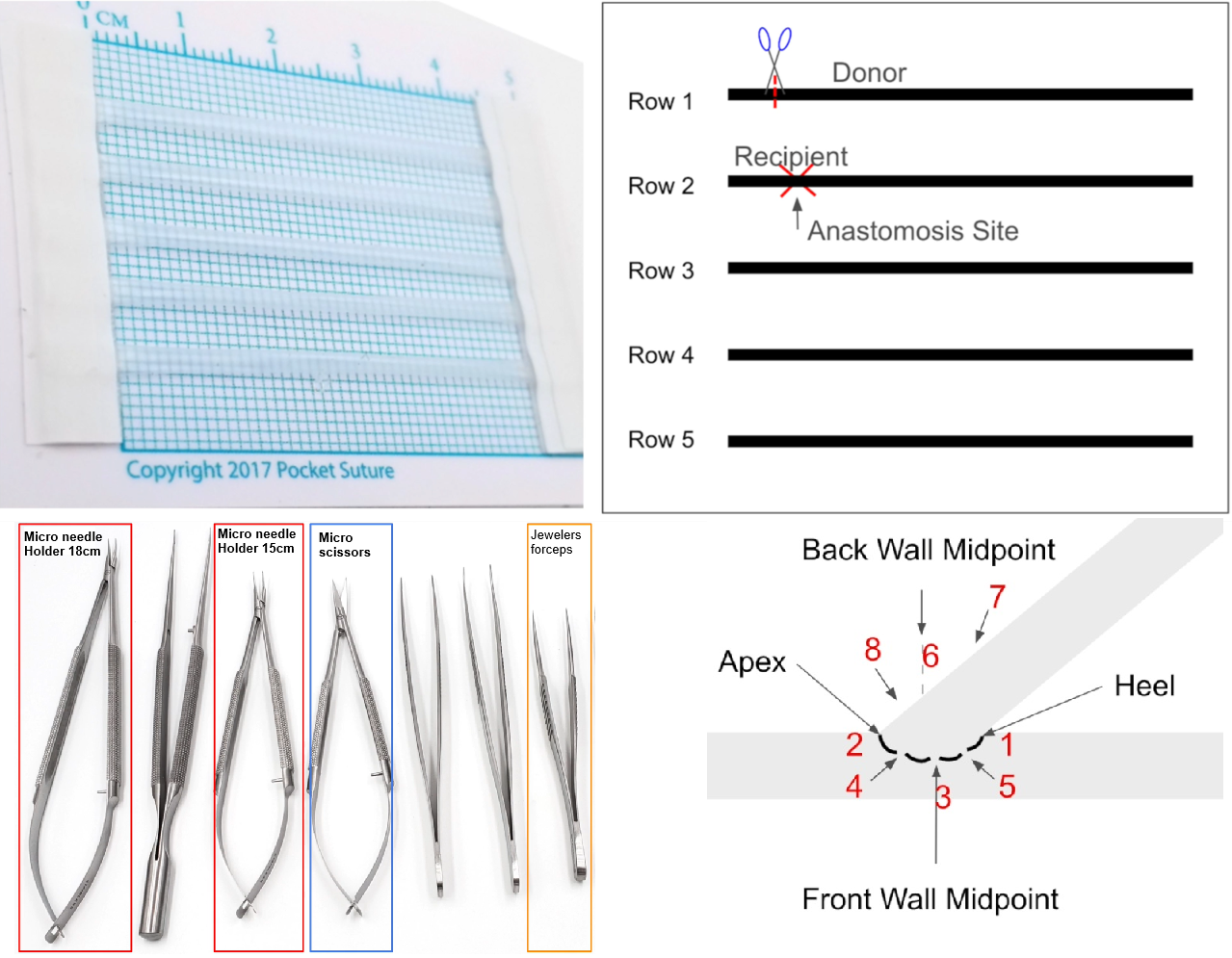} % Reduce the figure size so that it is slightly narrower than the column. Don't use precise values for figure width.This setup will avoid overfull boxes.
	\caption{Microvascular practice cards and the suturing stitches illustration.}
	\label{fig:pad_stitches}
\end{figure}

The standard practice procedure involves performing a complete end-to-side microanastomosis. This begins with an incision along the donor vessel, followed by trimming its tip to enlarge the lumen. A longitudinal cut is then made on the recipient vessel to accommodate the donor vessel. Anastomosis is performed using eight evenly spaced sutures. The suturing sequence followed a standardized order: starting with the heel (posterior right junction), followed by the apex (anterior left junction), the midpoint of the front wall, and the left and right sides adjacent to it. The final three sutures were placed at the midpoint of the back wall and on both side of this midpoint. The overall procedure and suture pattern are illustrated in Fig. \ref{fig:pad_stitches}.

\subsection{Transformer-based Action Segmentation}
\begin{figure}[!t]
	\centering
	%\fbox{\rule{0pt}{4in} \rule{0.9\linewidth}{0pt}}
	\includegraphics[width=\columnwidth]{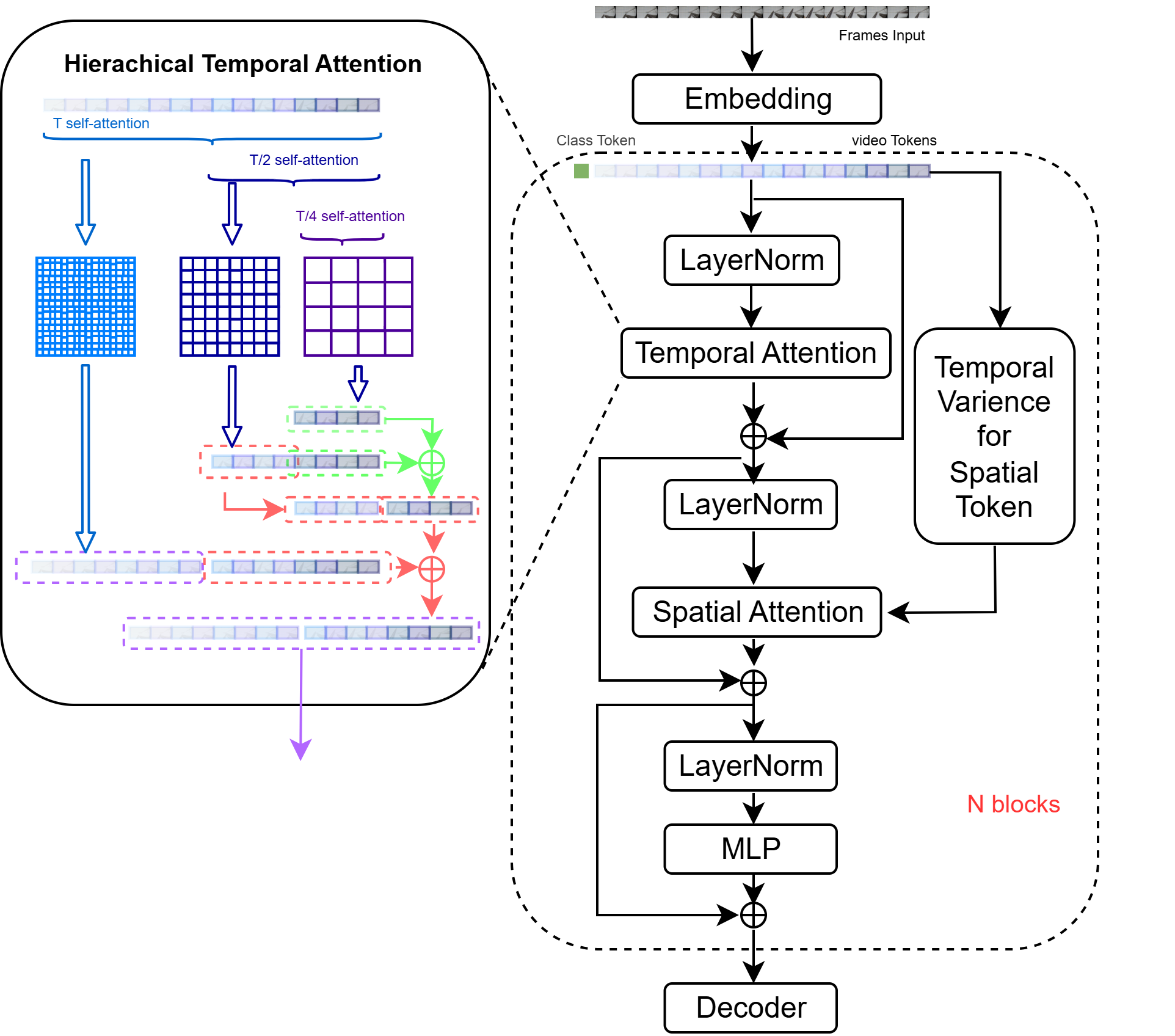}	
	\caption{The transformer architecture for microanastomosis action segmentation.}
	\label{fig:transformer}
\end{figure}

We developed a surgical action segmentation architecture building upon TimeSformer \cite{bertasius2021space} and Surgformer \cite{yang2024surgformer} to extract spatiotemporal features from surgical video sequences. The model is designed to capture both hierarchical temporal dependencies and spatial dynamics critical to surgical task understanding. It employs a spatiotemporal tokenization scheme that incorporates hierarchical temporal attention to model the temporal structure, alongside a spatial self-attention mechanism modulated by temporal variance to emphasize contextually informative regions. The overall architecture is illustrated in Fig. \ref{fig:transformer}.

\subsubsection{Hierarchical Temporal Attention}
To model multi-scale temporal dependencies, we apply a hierarchical attention structure that combines global and local temporal attention. Given a sequence of $T$ RGB frames $\{I_{1}, I_{2}, ..., I_{T}\}$, each frame $I_t \in \mathbb{R}^{H \times W \times C}$ at time $t$ is divided into non-overlapping patches of size $P \times P$, producing $K = H \times W / P^2$ patches per frame.  Each patch is flattened and projected into a token vector using a learnable linear embedding layer. A special class token is appended to the token sequence. Learnable positional encodings are added to retain spatial and temporal order. The full sequence of tokens for the entire clip is thus:

\begin{equation}
	\begin{split}
		% & is used to align equation
		& \mathbf{X} = \{[CLS], x_{1,1},...,x_{1,K},...,x_{T,1},...,x_{T,K} \}  \\
		& \mathbf{X} \in \mathbb{R}^{(1+T \cdot K) \times d}
	\end{split}
	\label{eq:spattiotemporal token}
\end{equation}

\subsubsection{Global Temporal Attention.}
For each spatial location $i$, the token sequence $\{x_{1,i}, x_{2,i},...,x_{T,i}\}$ represents the temporal evolution of that location. We apply self-attention across time T to capture global temporal dynamics. This is done in parallel for all spatial positions:

\begin{equation}
	Attn_i =SelfAttention(\{x_{1,i}, x_{2,i},...,x_{T,i}\})
	\label{eq:temporal attention}
\end{equation}

\subsubsection{Local Temporal Attention.}
To refine features around the target frame, we apply local attention in short temporal windows, for example $T/2$ and $T/4$ of the target frame. This captures fine-grained motion and local temporal consistency.

The outputs from global and local temporal attention block are aggregated to produce temporally contextualized tokens for each patch location.

\subsubsection{Variance-Weighted Spatial Self-Attention}
Following temporal modeling and layer normalization, a weighted self-attention mechanism is applied to enhance spatial feature representation. Instead of aggregating spatial tokens using uniform pooling or equal attention, we introduce a variance-guided spatial weighting mechanism that leverages temporal variability. Specifically, spatial tokens are reweighted prior to spatial self-attention based on the variance of their activations across time, assigning higher importance to locations that exhibit greater temporal dynamics. This reweighting modulates the input to the attention mechanism, indirectly emphasizing informative regions, while preserving standard spatial attention computation per frame.

For each spatial position $i$, we compute the temporal variance of its token representation across the sequence in Equation \ref{eq:varience}:
\begin{comment}
	\begin{equation}
		\begin{split}
			\mu_i &= \frac{1}{T} \sum_{t=1}^{T} x_{t,i} \\
			\sigma_i^2 &= \frac{1}{T} \sum_{t=1}^{T} \left\| x_{t,i} - \mu_i \right\|^2
		\end{split}
		\label{eq:varience}
	\end{equation}
\end{comment}
\begin{equation}
		\mu_i = \frac{1}{T} \sum_{t=1}^{T} x_{t,i}, \quad
		\sigma_i^2 = \frac{1}{T} \sum_{t=1}^{T} \left\| x_{t,i} - \mu_i \right\|^2
	\label{eq:varience}
\end{equation}

This variance is used to derive spatial importance weights, normalized across all spatial locations using a softmax function as in Equation \ref{eq: weight}. Each spatial token $x_{i,i}$ is multiplied by its corresponding importance weight $\tilde{x}_{i,i} = w_i \cdot x_{i,i}$. This reweighting is applied prior to spatial self-attention, allowing the attention mechanism to prioritize dynamic and informative regions during feature aggregation. 

\begin{equation}
	w_i = \frac{\exp(\sigma_i^2)}{\sum_j \exp(\sigma_j^2)}
	\label{eq: weight}
\end{equation}

The features are subsequently passed through several sequentially stacked transformer blocks, followed by a feed-forward classification head consisting of LayerNorm, dropout, and a multi-layer perceptron (MLP) to generate the final action class predictions.

\subsection{Kinematic Feature Extraction}
To obtain precise motion profiles of surgical instruments during microanastomosis procedures, we implemented a robust object tracking pipeline that integrates object detection and multi-object tracking using You Only Look Once (YOLO) and Deep Simple Online and Realtime Tracking (DeepSORT), respectively \cite{redmon2016you,wojke2017simple}. While these methods demonstrate strong performance in various video understanding tasks, their application to microsurgical procedures introduces unique challenges. In microanastomosis scenes, instrument motions appear disproportionately large and abrupt due to the high magnification, in contrast to the smoother and more predictable movements observed in the unmagnified real world. This exaggerated motion effect necessitates targeted adaptations to maintain temporal consistency and enable precise localization of instrument tips across long and complex video sequences. Direct application of conventional methods to microsurgical scenarios introduces three recurring challenges: (1) visually similar instruments with only partially visible segments in the magnified field often lead to frequent class label switching in YOLO; (2) abrupt movements and frequent occlusions result in missed detections by YOLO; and (3) DeepSORT tends to produce imprecise or drifting bounding boxes, particularly during sudden instrument motions or partial occlusions.

To address these issues and improve spatial-temporal coherence, we designed a dual identification guided detection correction mechanism.

\noindent\textbf{Detection refinement:} When a high-confidence YOLO detection overlaps with a DeepSORT-predicted bounding box, we prioritize the detection result to refine the bouding box location and shape, correcting any drift introduced by the DeepSORT tracker.

\noindent\textbf{Class-label anchoring:} Each object instance is associated with two parallel labels, a persistent object ID from DeepSORT and a class label from YOLO detection. we propagate the most recent correct label associated with the corresponding object ID, preventing cascading errors in YOLO classification.

\noindent\textbf{Reassignment of tracking identities:} When the tracker assigns a new object ID due to temporary occlusion or loss of detection, we retrospectively align it with its prior ID based on class, time gap, and appearance similarity to restore continuity. This minimizes fragmentation in the tracking stream and supports consistent feature extraction across extended action segments.

Following instrument tracking, instrument tip positions are extracted from the bounding boxes for subsequent motion analysis. Candidate keypoints are sampled along the convex hull of the instrument silhouette within each bounding box. Each candidate is then evaluated for tip likelihood using a predefined shape descriptor that encodes instrument geometric features. Cosine similarity in Equation \ref{eq:tip} is computed between each candidate point and the reference shape descriptor to identify the most likely tip location.

\begin{equation}	
	\label{eq:tip}
	\begin{aligned}
		\hat{p} = \arg\max_{i} \frac{ \mathbf{d}_{\text{ref}} \cdot \mathbf{d}_i }{ \|\mathbf{d}_{\text{ref}}\| \|\mathbf{d}_i\| }
	\end{aligned}
\end{equation}

\noindent where $\hat{p}$ is the instrument tip; $\mathbf{d}_{\text{ref}} \in \mathbb{R}^n$ is the reference object descriptor vector; $\mathbf{d}_i \in \{\mathbb{R}^n \}_{i=1}^{N}$ is a set of feature vectors from $N$ candidate points.

The resulting tip coordinates are mapped from local bounding box space to global coordinates in the frame, yielding a temporally coherent trajectory for each instrument. These trajectories are subsequently used to derive kinematic features such as velocity, acceleration, jerk and relative motions between instruments, which are used for downstream skill assessment tasks.

\subsection{Microanastomosis Skill Classification}
To objectively evaluate technical proficiency in microanastomosis procedures, we implement a supervised classification framework that predicts surgeon skill levels based on interpretable performance metrics. Following the NOMAT rubric, we assess five key aspects of performance: (1) overall instrument handling, (2) needle driving motion quality, (3) knot tying motion quality, (4) needle driving action-level performance, and (5) knot tying action-level performance. Each is graded on a five-point Likert scale.

The input features to the classification pipeline are extracted from two primary sources: instrument motion kinematics and action-level temporal statistics:
\begin{itemize}	
	\item \textbf{Kinematic features:} To characterize overall instrument handling and the quality of action-specific motion, we extract velocity, acceleration, and jerk for each individual instrument. In addition, we compute relative motion features between instruments, including inter-instrument distance, relative speed, and angular displacement.
	\item \textbf{Action statistics:} For critical actions, we include temporal features including the duration of each action instance, the number of repetitions, and the cumulative time spent performing each action type. These metrics capture both efficiency and consistency in task execution.
\end{itemize}

We employ the Gradient Boosting Classifier (GBC) for supervised skill classification due to its effectiveness in modeling non-linear decision boundaries, robustness to class imbalance, and its resistance to overfitting on small datasets \cite{konstantinov2021interpretable}.

\section{Experiments}
A comprehensive experimental study was conducted to assess the performance of the proposed AI-driven surgical skill evaluation framework. All procedures related to data acquisition and study implementation were reviewed and approved by the Institutional Review Board of the collaborating institution, ensuring adherence to established ethical guidelines for research involving human participants.

\begin{table*}[t!]
	\centering
	\begin{tabular}{|c|c|c|c|c|c|c|}
		\hline
		Method    &Action     & Accuracy     & Precision 	&Recall      & Jaccard  & F1 \\
		\hline
		\multirow{8}{*}{Our method}   & No	& -   	& 85.36 & 87.34	&75.52	&86.05 \\
		\cline{2-7}
		& vessel\_cutting & -	&86.11 &  89.06	&77.60	&87.38	\\
		\cline{2-7}
		& needle\_handling & -	&71.65 &  85.06	& 61.45 &  75.72\\
		\cline{2-7}
		&needle\_touch\_vessel & -	& 83.18 & 89.42	&75.78	&  86.01 \\
		\cline{2-7}
		& needle\_withdrawing & -	&84.16 &  67.83	& 62.21	& 74.92\\
		\cline{2-7}
		& knot\_tying 	& -	&85.25 &  76.30	& 67.02	&  80.21 \\
		\cline{2-7}
		& knot\_cutting & -	&87.31 &  92.88	&81.43	&89.74 \\
		\cline{2-7}
		& \multirow{4}{*}{full video}   & 87.66	&83.29 & 83.99	&71.57	& 82.86	\\
		\cline{1-1} \cline{3-7}
		Ours with post-process &  &93.62	&89.32	&88.71	&82.22	&88.32\\
		\cline{1-1} \cline{3-7}
		MS-TCN		&	&78.64	& 75.17	& 77.64	& 60.54	& 71.44	\\
		\cline{1-1} \cline{3-7}
		Surgformer	&	&82.47	&80.82	&79.54	&68.60	&75.96	\\
		\hline
	\end{tabular}
	\caption{The performance of the proposed action segmentation algorithms and comparison with SOTA methods. }
	\label{tab:segmentation_results}
\end{table*}

\subsection{Data Collection}
Nine medical practitioners (8 male, 1 female; mean age: 30.5 years) participated in this study, encompassing a wide range of microsurgical experience levels, from novice trainees to experienced neurosurgeons. All participants were predominantly right-handed, with one identifying as ambidextrous. Each session adhered to a standardized protocol as described in the method section, consisting of three vessel transections followed by eight sequential suture placements, allowing for consistent procedural structure and comparative analysis across recordings.We define six distinct and meaningful microanastomosis actions, while all remaining segments, either lacking specific actions or containing empty frames, are grouped under the label ``No''. A detailed description of the action categories is provided in Table \ref{tab:segmentation_results}. %\ref{tab:actions}.
\begin{comment}
	\begin{table*}[t!]
		\caption{Actions in a complete microanastomosis procedure. }
		\label{tab:actions}
		\centering
		\begin{tabular}{|c|c|c|c|c|c|c|c|}
			\hline
			Class ID    &0     & 1     & 2      & 3  & 4  & 5 & 6 \\
			\hline
			Action   & No	& vessel cutting   	&  needle handling   & needle touch vessel  	& needle withdrawing    & knot tying 	& knot cutting	\\
			\hline
		\end{tabular}
	\end{table*}
	
\end{comment}

Each participant completed between five and ten microanastomosis procedures under controlled laboratory conditions, yielding a total of 63 video recordings. Among these, 58 recordings captured the full procedure, with an average duration of approximately 26 minutes; the remaining five were either with wrong magnification or truncated due to technical issues or early termination.

All skill metrics in each video were independently assessed by two board-certified neurosurgeons using the NOMAT rubric. The evaluation results were reconciled through consensus vote. Technical performance was rated on a five-point Likert scale, where higher scores indicated greater proficiency. Due to the limited dataset size and class imbalance in the expert ratings, the original scores were discretized into three ordinal skill categories: Poor, Moderate, and Good, using thresholds at 2.5 and 3.5.

\subsection{Action Segmentation Performance}
\subsubsection{Model Training.} A total of 20 microanastomosis videos were manually annotated to train the transformer-based action segmentation model, with 15 videos used for training, 3 for validation, and 2 reserved for testing. The trained model was subsequently applied to segment surgical actions in the remaining 38 videos. All 58 videos were then utilized for downstream skill assessment.

The action segmentation model weights were initialized using parameters pretrained on the Kinetics dataset via TimeSformer \cite{kay2017kinetics}, while all task-specific layers were randomly initialized. Training was conducted for 50 epochs with a batch size of 16, using 4 NVIDIA T4 Tensor Core GPUs. Input videos were downsampled to 10fps and a temporal window of $T = 16$ frames for spatiotemporal tokenization and global temporal feature extraction. A multi-scale temporal representation was employed using three levels of detail ($T$, $T/2$, and $T/4$), combined with equal weighted averaging to enhance temporal features. The model was optimized using the AdamW optimizer with a learning rate of $9 \times 10^{-5} $ and a layer-wise learning rate decay factor of 0.75.

\subsubsection{Performance Metrics.} To evaluate action segmentation performance, we employ five widely-used benchmark metrics across both frame-level and action-level granularity. At the frame level, we report classification accuracy to assess the prediction result across videos. At the action level, we compute precision, recall, Jaccard index, and F1 score to quantify the correctness, completeness, and overall quality of predicted action segments relative to the ground truth. These metrics collectively provide a comprehensive assessment in the action segmentation results.

\subsubsection{Action Segmentation Results.} 
\begin{figure*}[t!]
	\centering
	\includegraphics[width=\textwidth]{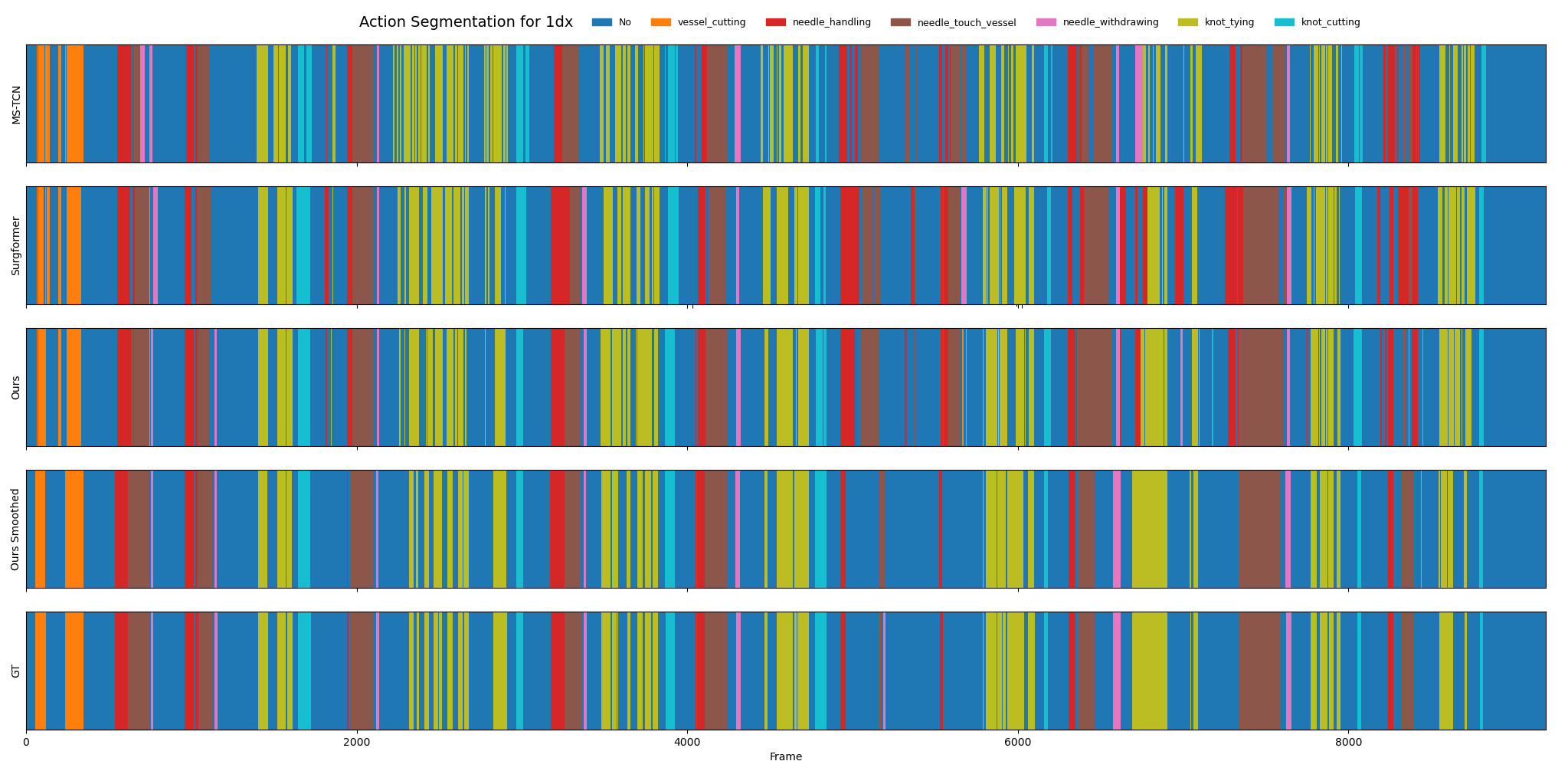} % Reduce the figure size so that it is slightly narrower than the column. Don't use precise values for figure width.This setup will avoid overfull boxes.
	\caption{Comparison of action segmentation results across different methods against the ground truth.}
	\label{fig:action_seg}
\end{figure*}

The performance metrics of the proposed surgical transformer model are presented in Table \ref{tab:segmentation_results}. To further enhance segmentation quality, we apply a temporal smoothing strategy that filters out predicted action segments shorter than five frames, and incorporate an action dictionary-based post-processing step to correct common misclassifications. These refinements result in improved segmentation consistency temporally and semantically. We also compare its overall performance with two baseline method: MS-TCN and Surgformer. The results demonstrate that our transformer-based model without post-processing, outperforms MS-TCN and Surgformer by 11\% and 6\% in accuracy, respectively. When post-processing is applied, the accuracy improvement increases to 19\% over MS-TCN and 13\% over Surgformer. A qualitative comparison of action segmentation results from each method, along with ground truth annotations, is illustrated in Fig. \ref{fig:action_seg}.

\begin{table}[t!]
	\centering
	\begin{tabular}{|c|c|c|c|c|}
		\hline
		NOMAT    				&Level     & Precision 	&Recall & F1 \\
		\hline
		\multirow{3}{*}{HSI}   & Good	&0.86      &0.92      &0.89  \\
		\cline{2-5}
		&Moderate 	&0.85     & 0.85     & 0.85\\
		\cline{2-5}
		& Poor	&0.83     &0.77     & 0.80 	\\
		\hline
		\multirow{3}{*}{KT}   & Good	&0.75      &0.82      &0.78\\
		\cline{2-5}
		&Moderate 	&0.54      &0.68      &0.60 \\
		\cline{2-5}
		& Poor	&0.64     & 0.41     & 0.50	\\
		\hline
		\multirow{3}{*}{NHC}   & Good	&0.84      &0.93      &0.88   \\
		\cline{2-5}
		&Moderate 	&0.66     & 0.89     & 0.76\\
		\cline{2-5}
		& Poor	&	0.73      &0.39      & 0.51	\\
		\hline
		\multirow{3}{*}{MEN}   & Good	&0.84      &1.00      & 0.91\\
		\cline{2-5}
		&Moderate 	&0.65     & 0.88     & 0.75 	\\
		\cline{2-5}
		& Poor	&0.73     & 0.34     & 0.47	\\
		\hline
		\multirow{3}{*}{MEKT}   & Good	&0.87      &1.00     & 0.93 \\
		\cline{2-5}
		&Moderate 	&0.78      &0.94     & 0.85 \\
		\cline{2-5}
		& Poor	&0.90      &0.58      &0.70 	\\
		\hline
	\end{tabular}
	\caption{Microanastomosis skill classification results with NOMAT. HSI is the overall handling of surgical instrument, KT and NHC are the knot tying and needle handling with care motion performance, MEN and MEKT are the microsurgical efficiency with needle and knot tying, respectively. }
	\label{tab:skill_results}
\end{table}

\subsection{Microanastomosis Skill Assessment Results}
\begin{figure}[!t]
	\centering
	\includegraphics[width=\columnwidth]{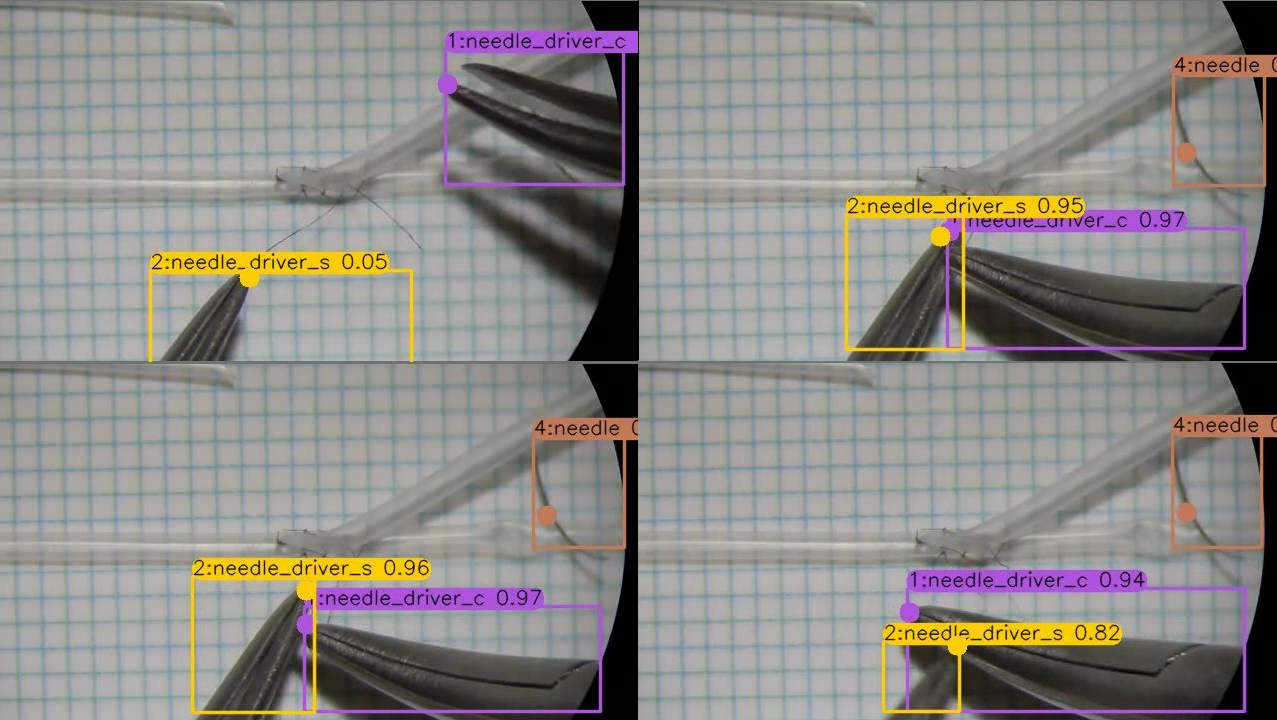} % Reduce the figure size so that it is slightly narrower than the column. Don't use precise values for figure width.This setup will avoid overfull boxes.
	\caption{YOLO-based instrument tip tracking.}
	\label{fig:tips}
\end{figure}

Action segmentation prediction was performed on the remaining 38 videos in the dataset. Video-level and action-level motion features were then extracted using the YOLO-based instrument tip localization method, the tracking result is illustrated in Fig. \ref{fig:tips}. These features served as input to supervised Gradient Boosting Classifiers, trained to replicate expert grading based on annotated score labels by experienced raters. The dataset was divided into 80\% for training and 20\% for testing. Within the training set, five-fold cross-validation was employed, wherein the model was iteratively trained on four folds and validated on the fifth to promote generalization and reduce overfitting. The model achieved an accuracy of 84.8\% for overall instrument handling, 63.4\% for knot tying motion performance, and 73.8\% for needle handling motion. For microsurgical action efficiency, the model attained 74.0\% accuracy in needle handling actions and 84.0\% in knot tying action. The average accuracy across all evaluated metrics was 76.0\%. The resulting classification performance for each skill level is summarized in Table \ref{tab:skill_results}. 

Given the limited dataset size and the involvement of only two expert raters for labeling, the resulting annotations may be subject to bias and limited objectivity. We anticipate that the performance of the machine learning classification model can be further improved through the inclusion of a larger dataset and a more diverse panel of surgeon raters in future work.

\section{Conclusion}
In this study, we introduce a novel framework for automated, action-level assessment of surgical proficiency in microanastomosis procedures, combining transformer-based video segmentation with kinematic analysis and interpretable performance metrics. By leveraging the capabilities of transformer architectures for fine-grained temporal segmentation, our approach enables precise identification of microsurgical gestures directly from operative video. This facilitates detailed, objective, and clinically meaningful evaluation of technical performance. Unlike prior approaches that focus primarily on recognizing what actions were performed, our method emphasizes how well each action was executed, aligning closely with structured NOMAT rubric.

The system delivers consistent, granular feedback, offering a scalable and interpretable alternative to conventional assessment methods. More importantly, the broader impact of this work lies in its accessibility and relevance to low- and middle-income countries, where barriers such as limited faculty availability, scarce simulation resources, and high training burdens continue to challenge surgical education. By enabling cost-effective, video-based assessment using online cloud service, our framework supports equitable access to surgical skill development and  high-quality feedback. This contributes to the global effort to enhance surgical training outcomes and improve the safety and quality of care in underserved settings.

\bibliography{ref}


\end{document}